\documentclass[10pt,twocolumn]{ieeeconf}
\usepackage{color,url}
\usepackage{float}
\usepackage{multirow}
\usepackage{amsmath}
\usepackage{amssymb,eqnarray}
\usepackage{mathabx}
\usepackage{graphicx}
\usepackage{stmaryrd}
\usepackage{psfrag}
\usepackage{pstool}

\makeatletter
\newcommand{\pushright}[1]{\ifmeasuring@#1\else\omit\hfill$\equationstyle#1$\fi\ignorespaces}
\newcommand{\pushleft}[1]{\ifmeasuring@#1\else\omit$\equationstyle#1$\hfill\fi\ignorespaces}
\makeatother

\usepackage{multicol}
\usepackage{bbm}
\usepackage[noend]{algorithmic}
\usepackage{algorithm2e}
\usepackage[english]{babel}
\usepackage{subfigure}
\usepackage{cite}
\usepackage{mathtools}
\mathtoolsset{showonlyrefs=false}
\makeatletter

\floatstyle{ruled}
\newfloat{algorithm}{tbp}{loa}
\providecommand{\algorithmname}{Algorithm}
\floatname{algorithm}{\protect\algorithmname}

\newtheorem{theorem}{\protect\theoremname}
\newtheorem{defn}{\protect\definitionname}
\newtheorem{proposition}{\protect\propositionname}

\newtheorem{assum}{\protect\assumname}
\newtheorem{problem}{\protect\probname}

\providecommand{\definitionname}{\textbf{Definition}}
\providecommand{\propositionname}{\textbf{Proposition}}
\providecommand{\remarkname}{\textbf{Remark}}
\providecommand{\theoremname}{\textbf{Theorem}}
\providecommand{\lemmaname}{Lemma}
\providecommand{\assumname}{\textbf{Assumption}}
\providecommand{\probname}{\textbf{Problem}}

\IEEEoverridecommandlockouts

\usepackage[margin=.7in]{geometry}
\title{Risk-Averse Planning Under Uncertainty}
\author{Mohamadreza Ahmadi, Masahiro Ono, Michel D. Ingham,\\ Richard M. Murray, and Aaron D. Ames \thanks{M. Ahmadi, R. M. Murray, and  A. D. Ames are with the California Institute of Technology, 1200 E. California Blvd., MC 104-44, Pasadena, CA 91125,  e-mail: (\{mrahmadi, murray, ames\}@caltech.edu). M. Ono and M. D. Ingham are with the NASA Jet Propulsion Laboratory, 4800 Oak Grove Dr, Pasadena, CA 91109, e-mail: (\{masahiro.ono, michel.d.ingham\}@jpl.nasa.gov)}}

\begin{document}

\maketitle

\begin{abstract}
We consider the problem of designing policies for partially observable Markov decision processes (POMDPs) with dynamic coherent risk objectives. Synthesizing risk-averse \textit{optimal} policies for POMDPs  requires infinite memory and thus undecidable. To overcome this difficulty, we propose a method based on bounded policy iteration for designing stochastic but finite state (memory) controllers, which takes advantage of standard convex optimization methods. Given a memory budget and optimality criterion,  the proposed method modifies the stochastic finite state controller leading to sub-optimal solutions with lower coherent risk.
\end{abstract}

\section{Introduction}

With the rise of autonomous systems being deployed in real-world settings, the associated risk that stems from unknown and unforeseen circumstances is correspondingly on the rise. In particular, in safety-critical scenarios, such as aerospace applications, decision making should account for risk. For example, spacecraft control technology relies heavily on a relatively large and highly skilled mission operations
team that generates detailed time-ordered and event-driven sequences of commands. This approach will
not be viable in the future with increasing number of missions and a desire to limit the operations team
and Deep Space Network (DSN) costs. Future spaceflight missions will be at large distances and light-
time delays from Earth, requiring novel capabilities for astronaut crews and ground operators to manage
spacecraft consumables such as power, water, propellant, and life support systems to prevent mission failure.
In order to maximize the science returns under these conditions, the ability to deal with emergencies and
safely explore remote regions are becoming more and more important~\cite{mcghan2016resilient}. Even in Mars rover navigation problems, finding planning policies that minimize risk is of utmost importance due to the uncertainties present in Mars surface data~\cite{ono2018mars} as illustrated in Figure~1.

Risk can be quantified in numerous ways. For example, mission risks can be mathematically characterized in terms of chance constraints~\cite{ono2013probabilistic,ono2012closed,ono2015chance}. The preference of one risk measure over
another depends on factors such as sensitivity to rare events, ease of estimation from data, and computational tractability. Artzner \textit{et. al.}~\cite{artzner1999coherent} characterized a set of natural properties that are desirable for a risk measure, called a coherent risk measure, and  have henceforth obtained widespread
acceptance in finance and operations research, among others. An important example of a coherent risk measure is the conditional value-at-risk (CVaR) that has received significant attention in decision making problems such as Markov decision processes (MDPs)~\cite{chow2015risk,chow2014algorithms,prashanth2014policy,bauerle2011markov}. General coherent risk measures for MDPs were studied in~\cite{ruszczynski2010risk}, wherein it was further assumed the risk measure is \emph{time consistent}, similar to the dynamic programming property. Following the footsteps of the latter contribution, ~\cite{tamar2016sequential} proposed a sampling-based algorithm for MDPs with static and dynamic coherent risk measures using policy gradient and actor-critic methods, respectively (also, see a model predictive control technique for linear dynamical systems with coherent risk objectives~\cite{singh2018framework}). 

\begin{figure}[t] \centering{
\includegraphics[scale=.47]{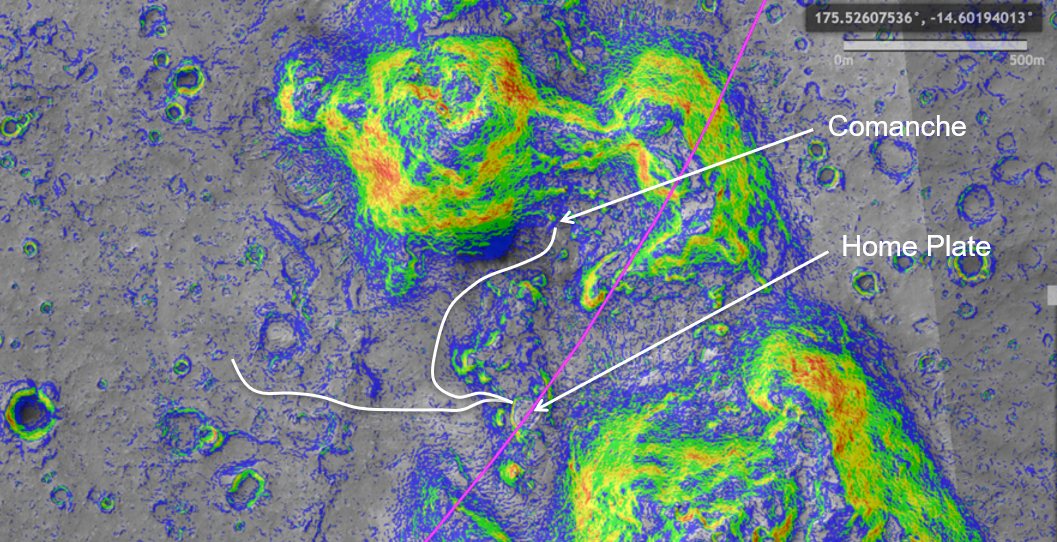}
\caption{Mars surface slope uncertainty for Mars rover navigation: regions with slopes ranging any values within (blue) $5^\circ-10^\circ$, (green) $10^\circ-15^\circ$, (yellow) $15^\circ-20^\circ$, (orange) $20^\circ-25^\circ$, (red)  $\ge 25^\circ$, and (the rest) $<5^\circ$ or no data. 
} }\label{fig:marssurface}
 \end{figure}
However, in many aerospace applications, sensing constraints does not allow for full-state observation and decision making involves partial observation~\cite{ahmadi2019safe,nilsson2018toward}. These problems can be represented as a partially observable Markov decision process (POMDP), where decision making is subject to uncertainty stemming from stochastic outcomes as well as partial observation~\cite{krishnamurthy2016partially}. In this paper, we propose a method based on bounded policy iteration to design sub-optimal risk-averse policies for POMDPs. To this end, we first discuss that the problem of designing risk-averse optimal policies is undecidable in general.  Then, we show that a stochastic but finite-memory controller can be synthesized to upper-bound the dynamic risk. Given a memory budget, we propose a policy iteration method to synthesize these finite-state controllers that can increase the number of memory states to improve risk-aversity. We illustrate our proposed method with a numerical example of path planning under uncertainty.

The rest of the paper is organized as follows. The next section reviews some preliminary notions and definitions used in the sequel. In Section~\ref{sec:riskpomdpfull}, we discuss POMDPs with coherent risk measures. In Section~\ref{sec:risksfc}, we propose sub-optimal stochastic finite state controllers that minimize the upper-bound on the coherent risk. In Section~\ref{sec:bpi}, a bounded policy iteration algorithm is formulated to design risk-averse stochastic finite controllers. In Section~\ref{sec:example}, we elucidate our results with a numerical example. Finally, in Section~\ref{sec:conclusions}, we conclude the paper and give directions for future research.

\section{Preliminaries}

In this section, we briefly review some notions and definitions used throughout the paper.

\subsection{Markov Chains}\label{sec:markovchains}


A Markov chain $\mathcal{M}$ is composed of the state space
$\mathcal{S}$, the transition probability defined as the conditional distribution $T(.|s):\mathcal{S}
\to [0,1]$ such that $\sum_{s'\in\mathcal{S}}T(s'|s)=1,\ \forall s\in\mathcal{S}$,
and the initial distribution $\iota_{\mathrm{init}}$ such that $\sum_{s\in\mathcal{S}}\iota_{\mathrm{init}}(s) = 1$. An infinite path, denoted by the superscript $\omega$, of the Markov chain $\mathcal{M}$ is a sequence of states $\pi = s_0s_1\dots \in \mathcal{S}^\omega$ such that $T(s_{t+1}|s_t)>0$ for all $t$ and $\iota_{\mathrm{init}}(s_0)>0$. The probability space over such paths is the defined as follows. The sample space $\Omega$ is the set of infinite paths with initial state $s \in \mathcal{S}$, \textit{i.e.}, $\Omega = \mathrm{Paths}(s)$. $\Sigma_{\mathrm{Paths}(s)}$ is the least $\sigma$-algebra on $\mathrm{Paths}(s)$ containing $\mathrm{Cyl}(\omega)$, where $\mathrm{Cyl}(\omega)=\{ \omega' \in \mathrm{Paths}(s) \mid \omega~\text{is a prefix of}~\omega'\}$ is the cylinder set. Finally, in order to specify the probability measure over all  sets of events in $\Sigma_{\mathrm{Paths}(s)}$, it is sufficient to provide the probability of each cylinder set, which can be computed as
 $
\mathrm{Pr}_{\mathcal{M}}\left[\mathrm{Cyl}(s_0\ldots s_n) \right]=\iota_{\mathrm{init}}(s_0) \prod_{0 \le t \le n}T(s_{t+1}\mid s_t).
$ 
Once the probability measure is defined over the cylinder sets, the expectation operator $\mathbb{E}_{\mathcal{M}}$ is also uniquely defined. In the sequel, we remove the subscript whenever the Markov chain is clear from the context.



\subsection{Partially Observable Markov Decision Process} \label{sec:POMDP}

\begin{defn}[POMDP] \label{defn:POMDP} {
A \emph{POMDP}, $\mathcal{PM}$, consists of:
\begin{itemize}
\item States $\mathcal{S}  =
\{s_{1} ,\dots,s_{|\mathcal{S}|} \}$ of the
autonomous agent(s) and world model,
\item Actions $Act = \{\alpha_{1},\dots,\alpha_{|Act|}\}$ available to the robot,
\item Observations $\mathcal{O} = \{o_{1},\dots,o_{|\mathcal{O}|}\}$,
\item A Transition function $T(s_{j} |s_{i} ,\alpha)$,
\item A cost, $c(s_{i},\alpha_i ) \ge 0$, for each state $s_{i}  \in \mathcal{S} $ and action~$\alpha_i \in {Act}$.
\end{itemize}
}
\end{defn}
\vspace{0.2cm}
This paper considers {\em finite} POMDPs where $\mathcal{S} $, $Act$, and $\mathcal{O}$
are finite sets. For each action the probability of making a transition from state $s_{i}  \in
\mathcal{S} $ to state $s_{j}  \in \mathcal{S} $ under action $\alpha \in Act$ is given by
$T(s_{j} |s_{i} ,\alpha)$. For each state $s_{i} $, an observation $o \in
\mathcal{O}$ is generated independently with probability $O(o|s_{i} )$. The starting world
state is given by the distribution $\iota_{\mathrm{init}}(s_i )$. The probabilistic components of a
POMDP model must satisfy the following:
\begin{equation*}
    \begin{cases}
    \sum_{s  \in \mathcal{S} } T(s |s_{i} ,\alpha) = 1, & \forall s_i  \in \mathcal{S} ,\alpha \in Act \\
    \sum_{o \in \mathcal{O}} O(o|s ) = 1, & \forall s  \in \mathcal{S} \\
    \sum_{s  \in \mathcal{S} } \iota_{\mathrm{init}}(s ) = 1. & {}
    \end{cases}
\end{equation*}

Given a POMDP, we can define beliefs or distributions over states at each time step to keep track of sufficient statistics with finite description~\cite{astrom65}. The beliefs $b \in \Delta(\mathcal{S})$, with $\Delta(\mathcal{S})$ being the set of probability distributions over $\mathcal{S}$, for all $s \in \mathcal{S}$ can be computed using the Bayes' law as follows:
\begin{align}
    b_0(s) &= \frac{\iota_{\mathrm{init}}(s)O(o_0\mid s)}{\sum_{o \in O} \iota_{\mathrm{init}}(s) O(o \mid s)},\\ \label{eq:beliefupdate}
    b_t(s) &= \frac{O(o_t \mid s,\alpha_t)\sum_{s' \in \mathcal{S}} T(s \mid s',\alpha_t)b_{N-1}(s')}{\sum_{s \in \mathcal{S}} O(o_t \mid s,\alpha_t)\sum_{s' \in \mathcal{S}} T(s \mid s',\alpha_t)b_{N-1}(s')},
\end{align}
for all $t\ge 1$. It is also worth mentioning that~\eqref{eq:beliefupdate} is referred to as the \emph{belief update equation}.

\subsection{Stochastic Finite State Control of POMDPs} \label{sec:FSC}


It is well established that designing optimal policies for POMDPs based on the (continuous) belief states require uncountably infinite memory or
internal states \cite{CassandraKL94, KLC98, MADANI20035}. This paper focuses on a particular class of POMDP controllers, namely, {\em stochastic finite state
controllers}. These controllers lead to a finite state space Markov chain for the closed loop
controlled system.
\vspace{0.2cm}
\begin{defn}[Stochastic Finite State Controller] \label{defn:sto-FSC} {
Let $\mathcal{PM}$ be a POMDP with observations $\mathcal{O}$, actions $Act$, and initial
distribution $\iota_{\mathrm{init}}$. A \emph{stochastic finite state controller } for
$\mathcal{PM}$ is given by the tuple $\mathcal{G} = (G,\omega,\kappa)$ where
%
\begin{itemize}
\item $G = \{g_1,g_2,\dots,g_{|G|}\}$ is a finite set of internal states~(I-states).
\item $\omega:G \times \mathcal{O} \to \Delta({G \times Act})$ is a function of internal stochastic finite state controller states  $g_k$ and observation $o$, such that $\omega(g_k,o)$ is a probability distribution over $G \times
Act$. The next internal state and action pair $(g_l,\alpha)$ is chosen by independent sampling of
$\omega(g_k,o)$. By abuse of notation, $\omega(g_l,\alpha|g_k,o)$ will denote the probability of
transitioning to internal stochastic finite state controller state $g_l$ and taking action $\alpha$, when the current internal
state is $g_k$ and observation $o$ is received.
\item $\kappa:\Delta({\mathcal{S}}) \to \Delta(G)$ chooses the starting internal FSC state $g_0$, by independent
sampling of $\kappa(\iota_{\mathrm{init}})$, given initial distribution $\iota_{\mathrm{init}}$ of $\mathcal{PM}$.
$\kappa(g|\iota_{\mathrm{init}})$ will denote the probability of starting the FSC in internal state $g$ when
the initial POMDP distribution is $\iota_{\mathrm{init}}$.
\end{itemize}
}
\end{defn}
\vspace{0.2cm}
Closing the loop around a POMDP with a stochastic finite state controller yields the
following transition system.
\vspace{0.2cm}
\begin{defn}[Global Markov Chain] \label{defn:globalMC} {
Let POMDP $\mathcal{PM}$ have state space $\mathcal{S}$ and let $G$ be the I-states of stochastic finite state controller 
$\mathcal{G}$. The global Markov chain $\mathcal{M}^{\mathcal{PM},\mathcal{G}}_{\mathcal{S}\times
G}$ (or simply $\mathcal{M}$, where the stochastic finite state controller and the POMDP are clear from the context) with execution $\sigma = \lbrace[s_0,g_0],[s_1,g_1],\dots\rbrace,\ [s_t,\ g_t] \in \mathcal{S}
\times G$ evolves as follows:
\begin{itemize}
\item The probability of initial global state $[s_0,g_0]$ is
  \begin{equation*} \label{eq:GlobalMCInitial}
     \iota_{\mathrm{init}}^{\mathcal{M}}\left(\left[s_0,g_0 \right]\right) 
            = \iota_{\mathrm{init}}(s_0)\kappa(g_0|\iota_{\mathrm{init}}).
  \end{equation*}
\item The state transition probability, $T^{\mathcal{M}}$, is given by
  \begin{equation*} \label{eq:GlobalMCTransition}
    \begin{aligned}
      T^{\mathcal{M}} & \left(\left[s_{t+1},g_{t+1}\right] \left|
            \left[s_t,g_t\right] \right. \right)  = \\ 
       \sum_{o\in\mathcal{O}} &
            \sum_{\alpha \in Act}O(o|s_t)\omega(g_{t+1},\alpha |g_t,o)T(s_{t+1}|s_t,\alpha).
    \end{aligned}
  \end{equation*}
\end{itemize}
}
\end{defn}
\vspace{0.2cm}
Note that the global Markov chain arising from a finite state space POMDP also has a finite state
space.

\subsection{Coherent Risk Measures}

Consider a probability space $(\Omega, \mathcal{F}, P)$, a filteration $\mathcal{F}_0 \subset \cdots \mathcal{F}_N \subset \mathcal{F} $, and an adapted sequence of random variables~(stage-wise costs) $c_t,~t=0,\ldots, N$, where $N \in \mathbb{N}_{\ge 0} \cup \{\infty\}$.
We further define the spaces $\mathcal{C}_t = \mathcal{L}_p(\Omega, \mathcal{F}_t, P)$, $p \in [0,\infty)$, $t=0,\ldots,N$ and let $\mathcal{C}_{t:N}=\mathcal{C}_t\times \cdots \times \mathcal{C}_N$ and $\mathcal{C}=\mathcal{C}_0\times \mathcal{C}_1 \times \cdots$. We further assume that the sequence $c \in \mathcal{C}$ is almost surely bounded, \textit{i.e.}, 
$$
\max_t \mathrm{essup}~| c_t(\omega) | < \infty.
$$

In order to describe how one can evaluate the risk of subsequence $c_t,\ldots, c_N$ from the perspective of stage $t$, we require the following definitions.

\vspace{0.2cm}
\begin{defn}[Conditional Risk Measure]{
A mapping $\rho_{t:N}: \mathcal{C}_{t:N} \to \mathcal{C}_{t}$, where $0\le t\le N$, is called a \emph{conditional risk measure}, if it has the following monoticity property:
\begin{equation}
    \rho_{t:N}(c) \le   \rho_{t:N}(c'), \quad \forall c, \forall c' \in \mathcal{C}_{t:N}~\text{such that}~c \le c',
\end{equation}
where the inequalities should be understood componentwise.
}
\end{defn}
\vspace{0.2cm}
\begin{defn}[Dynamic Risk Measure]
{A \emph{dynamic risk measure} is a sequence of conditional risk measures $\rho_{t:N}:\mathcal{C}_{t:N}\to \mathcal{C}_{t}$, $t=0,\ldots,N$.}
\end{defn}
\vspace{0.2cm}
One fundamental property of dynamic risk measures is their consistency over time. That is, if $c$ will be as good as $c'$ from the perspective of some future time $\theta$, and they are identical between time $\tau$ and $\theta$, then $c$ should not be worse than $c'$ from the current time's perspective.
\vspace{0.2cm}
\begin{defn}[Time-Consistent Risk Measure]{
A dynamic risk measure $\left\{ \rho_{t:N}  \right\}_{t=0}^T$ is called \emph{time-consistent} if for all $0\le t \le \tau < \theta  \le T$ and all sequences $Z,W \in \mathcal{C}_{t:N}$ the conditions
\begin{multline*}
c_t =c'_t,~t=\tau,\ldots,\theta-1,~~\text{and}~\\ \rho_{\theta,T}(Z_\theta,\ldots,c_t) \le  \rho_{\theta,T}(W_\theta,\ldots,c'_t)
\end{multline*}
imply
\begin{equation}
\rho_{\tau,N}(c_\tau,\ldots,c_t) \le  \rho_{\tau,N}(c'_\tau,\ldots,c'_t).
\end{equation}
}
\end{defn}
\vspace{0.2cm}
If a risk measure is time-consistent, we can define the one-step conditional risk measure $\rho_t:\mathcal{C}_{t+1}\to \mathcal{C}_t$, $t=0,\ldots,N-1$ as follows:
\begin{equation}
    \rho_t(c_{t+1}) = \rho_{t,t+1}(0,c_{t+1}),
\end{equation}
and for all $t=1,\ldots,N$, we obtain:
\begin{multline}
    \label{eq:dynriskmeasure}
    \rho_{t,N}(c_t,\ldots,c_N)= \rho_t \big(c_t + \rho_{t+1} ( c_{t+1}+\rho_{t+2}(c_{t+2}+\cdots\\
    +\rho_{N-1}\left(c_{N-1}+\rho_{N}(c_N) \right) \cdots )) \big).
\end{multline}
Note that the time-consistent risk measure is completely defined by one-step conditional risk measures $\rho_t$, $t=0,\ldots,N-1$ and, in particular, for $t=0$, \eqref{eq:dynriskmeasure} define a risk measure of the entire sequence $c \in \mathcal{C}_{0:N}$.

At this point, we are ready to define a coherent risk measure. 
\vspace{0.2cm}
\begin{defn}[Coherent Risk Measure]\label{defi:coherent}{
We call the one-step conditional risk measures $\rho_t: \mathcal{C}_{t+1}\to \mathcal{C}_t$, $t=1,\ldots,N-1$ as in~\eqref{eq:dynriskmeasure} a \emph{coherent risk measure} if it satisfies the following conditions
\begin{itemize}
    \item \textbf{Convexity:} $\rho_t(\lambda c + (1-\lambda)c') \le \lambda \rho_t(c)+(1-\lambda)\rho_t(c')$, for all $\lambda \in (0,1)$ and all $c,c' \in \mathcal{C}_{t+1}$;
    \item \textbf{Monotonicity:} If $c\le c'$ then $\rho_t(c) \le \rho_t(c')$ for all $c,c' \in \mathcal{C}_{t+1}$;
    \item \textbf{Time Consistency:} $\rho_t(c+c')=c+\rho_t(c')$ for all $c \in \mathcal{C}_t$ and $c' \in \mathcal{C}_{t+1}$;
    \item \textbf{Positive Homogeneity:} $\rho_t(\beta c)= \beta \rho_t(c)$ for all $c \in \mathcal{C}_{t+1}$ and $\beta \ge 0$.
\end{itemize}
}
\end{defn}
\vspace{0.2cm}


Henceforth, all the risk measures considered are assumed to be coherent. In this paper, we are interested in the discounted infinite horizon problems. Let $\gamma \in (0,1)$ be a given discount factor. For $N=0,1,\ldots$, we define the functionals 
\begin{multline}
    \rho^\gamma_{0,N}(c_0,\ldots,c_N) = \rho_{0,N}(c_0,\gamma c_1,\ldots, \gamma^{N}c_N) \nonumber \\
                                      = \rho_0 \bigg(c_0 + \rho_{1} \big( \gamma c_{1}+\rho_{2}(\gamma^2c_{2}+\cdots \nonumber\\
                                      ~~+\rho_{N-1}\left(\gamma^{N-1}c_{N-1}+\rho_{N}(\gamma^{N}c_N) \right) \cdots )\big) \bigg),
\end{multline}
which are the same as~\eqref{eq:dynriskmeasure} for $t=0$, but with discounting $\gamma^{t}$ applied to each $c_t$. Finally, we have total discounted risk functional $\xi_{\gamma}:\mathcal{C}\to \mathbb{R}$ defined as \begin{equation}\label{eq:totaldiscrisk} \xi_{\gamma}(Z) = \lim_{N \to \infty} \rho^\gamma_{0,N}(c_0,\ldots,c_N).\end{equation} From~\cite[Theorem 3]{ruszczynski2010risk}, we have that $\xi_{\gamma}$ is convex, monotone, and positive homogenoeus. 


\section{Risk-Averse POMDPs} \label{sec:riskpomdpfull}

Notions of coherent risk and dynamic risk measures discussed in the previous section have been developed and applied in microeconomics and mathematical finance fields in the past two decades~\cite{vose2008risk}. Generally speaking, risk-averse decision making is concerned with the behavior of agents, e.g. consumers and investors, who, when exposed to uncertainty, attempt to lower that uncertainty. The agent averts to agree to a situation with an unknown payoff rather than another situation with a more predictable payoff but possibly lower expected payoff. In a Markov decision making setting, the main idea in risk-averse control is to replace the conventional conditional expectation of the cumulative reward or cost objectives with more general risk measures. 

Consider a stationary (policies, transition probabilities, and cost functions do not depend explicitly on time) controlled Markov process $\{s_t\}$, $t=0,1,\ldots$. Each policy $\pi = \{\pi_t\}_{t=0}^\infty$ leads to a cost sequence $c_t=c(s_t,\alpha_t)$, $t=0,1,\ldots$. We define the dynamic risk of evaluating the $\gamma$-discounted cost of the policy $\pi$ as
\begin{equation}\label{eq:objrisk}
    V_\gamma(\pi,s_0) = \xi_{\gamma} \big( c(s_0,\alpha_0),c(s_1,\alpha_1),\ldots \big),
\end{equation}
where $\xi_{\gamma}$ is defined in~\eqref{eq:totaldiscrisk}. In this work, we are interested in addressing the following problem:
\vspace{0.2cm}
\begin{problem}\textit{
For a given POMDP~$\mathcal{PM}$, a discount factor $\gamma \in (0,1)$, and a total risk  functional $V_\gamma$ as in~\eqref{eq:objrisk} with $\{\rho_t\}_{t=0}^\infty$ being coherent risk measures, compute 
$$
\pi^* \in \mathrm{argmin}_{\pi} V_\gamma(\pi,b_0).
$$
}
\end{problem}
\vspace{0.2cm}
We refer to a controlled Markov process with the ``nested'' objective~\eqref{eq:objrisk} a \emph{risk-averse} Markov process. Many applications such as portfolio allocation problems~\cite{gonzalo2019differences} and organ transplant decisions~\cite{heilman2017potential} require a risk-averse Markov model. It was also previously demonstrated in~\cite{chow2015risk,osogami2012robustness} that coherent risk measure objectives can account for  modeling errors and parametric uncertainty in MDPs. 

The main challenge is that at any time $t$, the value of $\rho_t$ is $\mathcal{F}_t$-measurable and is allowed to depend on the entire history of the process $\{s_0,s_1,\ldots\}$ and we cannot expect to obtain a Markov optimal policy~\cite{ott2010markov}. 


In order to obtain Markov optimal policies, we need to make the following assumption (see~\cite[Section 4]{ruszczynski2010risk} for more details):
\vspace{0.2cm}
\begin{assum}\label{assum1}\textit{
For any function $\phi(s_t,a_t,s_{t+1})$, we have
\begin{equation}\label{eq:markovtrans}
    \rho_t(\phi(s_t,a_t,s_{t+1})) = \mathrm{R}\left\{\phi(s_t,a_t,\cdot),s_t,p(s_t,a_t) \right\}, 
\end{equation}
where $a_t=\pi(s_t)$. The function $\mathrm{R}$ is called a \emph{Markov risk transition mapping}.
}
\end{assum}
\vspace{0.2cm}
Note that the Markov risk transition mapping depends on the function $\phi$, the states $s$, and probability vector $p(s,a)$. The dot in $\phi(s_t,a_t, \cdot)$ on the right hand side of~\eqref{eq:markovtrans} represents a dummy variable that is integrated/summed out with respect to the $s_t$-th row of the transition probability matrix $p(s_t,a_t)$. The simplest case of the Markov risk transition mapping is the conditional expectation $\mathbb{E}\{\cdot \mid s_t,a_t\}$, \textit{i.e.}, 
\begin{multline*}
\mathrm{R}\left\{\phi(s_t,a_t,\cdot),s_t,p(s_t,a_t) \right\} = \mathbb{E}\{ \phi(s_t,a_t,s_{t+1})\mid s_t,a_t\} \\ = \sum_{s_{t+1}} \phi(s_t,a_t,s_{t+1}) T(s_{t+1}\mid s_t,a_t).
\end{multline*}
If $R$ is a coherent risk measure as described in Definition~\ref{defi:coherent}, then the Markov policies are sufficient to ensure optimality~\cite{ruszczynski2010risk}. In particular, for the CVaR risk measure, the Markov risk transition mapping is given by
\begin{multline}
    \mathrm{R} \{ \phi, s, p(s,a) \} \\= \inf_{z\in \mathbb{R}} \left\{ z + \frac{1}{\alpha} \sum_{s'}\left(\phi(s,a,s')-z\right)_{+} T(s'\mid s,a)        \right\}.
\end{multline}

The risk-averse formulation can be extended to POMDPs as follows.  
\vspace{0.2cm}
\begin{theorem}\textit{
Consider the POMDP $\mathcal{PM}$  with the nested risk objective~\eqref{eq:objrisk} and $\gamma \in (0,1)$. Let Assumption~\ref{assum1} hold, let $\rho_t,~t=0,1,\ldots$ be  coherent risk measures as described in Definition~\ref{defi:coherent}, and suppose $c(\cdot,\cdot)$ be non-negative and upper-bounded. Then, the stationary optimal policy $\pi^* = \{\pi^*_t\}_{t=0}^\infty = \{\pi^*\}_{t=0}^\infty$ is the solution of the following Bellman's equations
\begin{subequations}\label{eq:bellmanbeliefmdp}
\begin{align}
        V_\gamma(b) &= \min_{\alpha \in Act} \Big(c(b,\alpha) + \gamma \mathrm{R}\left \{ V_\gamma(b'),b, p(b' \mid b,\alpha) \right\} \Big), \\
        \pi^*(b) &\in  \mathrm{argmin}_{\alpha \in Act} \Big(c(b,\alpha) + \gamma \mathrm{R}\left \{ V_\gamma(b'),b, p(b' \mid b,\alpha) \right\} \Big),
\end{align}
\end{subequations}
where $c(b,\alpha) = \sum_{s \in \mathcal{S}} c(s,\alpha)b(s)$.
}
\end{theorem}
\vspace{0.2cm}
\begin{proof}
Note that a POMDP can be represented as an MDP over the belief states~\eqref{eq:beliefupdate}. Hence, a POMDP is a controlled Markov process with states $b \in \Delta(\mathcal{S})$, where the controlled belief transition probability is described as
\begin{multline}\label{eq:belieftransition}
 p(b' \mid b,\alpha) = \sum_{o \in \mathcal{O}} p(b' \mid b, o, \alpha)~p(o \mid b, \alpha)  \\= \sum_{o \in \mathcal{O}}\delta \left(b' - \frac{O(o \mid s,\alpha)\sum_{s' \in \mathcal{S}} T(s \mid s',\alpha)b(s')}{\sum_{s \in \mathcal{S}} O(o \mid s,\alpha)\sum_{s' \in \mathcal{S}} T(s \mid s',\alpha)b(s')}\right)  \\ \times \sum_{s \in \mathcal{S}} O(o \mid s,\alpha)\sum_{s'' \in \mathcal{S}} T(s \mid s'',\alpha)b(s''),
\end{multline}
with $$\delta(a) = \begin{cases} 1 & a=0, \\ 0 & \text{otherwise}. \end{cases}$$  Then, given that $c(\cdot,\cdot)$ is non-negative and upper-bounded, from~\cite[Theorem 8.6.2]{krishnamurthy2016partially} and~\cite[Theorem 4]{ruszczynski2010risk}, we infer that from the Bellman equations~\eqref{eq:bellmanbeliefmdp} we can obtain the optimal policies. 
\end{proof}

We can use a method based on policy iteration to solve the dynamic programming equations~\eqref{eq:bellmanbeliefmdp} to design risk-averse optimal policies. To this end, for $k=0,1,\ldots$, given a stationary Markov policy $\pi^k$, we calculate the corresponding value function as
\begin{subequations}
\begin{equation}
    V^k_\gamma(b) = c\left(b,\pi^k(b)\right) + \gamma \mathrm{R}\left \{ V^k_\gamma(b), b, p(b'\mid b, \pi^k(b)       \right\}.
\end{equation}
Then, we compute the next policy $\pi^{k+1}$ as
\begin{equation}
    \pi^{k+1}(b) \in \mathrm{argmin}_\pi \Big(   c(b, \pi(b)) + \gamma        \mathrm{R}\big \{ V^k_\gamma(b), b, p(b'\mid b, \pi(b)       \big\}\Big).
\end{equation}
\end{subequations}

Unfortunately, the problem of designing risk-averse optimal Markovian policies for POMDPs is undecidable in general. This follows from~\cite[Theorem 4.4]{MADANI20035} by noting that $\inf_\pi V_\gamma = \sup_\pi \left(-V_\gamma \right)$.


In the subsequent section, we demonstrate that, if instead of considering policies with infinite-memory we search over finite-memory policies, then we can minimize upper-bounds on the total risk cost functional~\eqref{eq:objrisk}. 




\section{Risk-Averse Stochastic Finite State Controllers}\label{sec:risksfc}

Under a stochastic finite state controller, the POMDP is transformed into a Markov chain $\mathcal{M}^{\mathcal{PM} \times \mathcal{G}}_{\mathcal{S} \times \mathcal{G}}$ with design probability distributions $\omega$ and $\kappa$. We define the total risk functional of this parametric Markov chain as
\begin{equation}\label{eq:objriskfsc}
    V_\gamma(\mathcal{G},\iota_{\mathrm{init}}) = \xi_{\gamma} \big( c([s_1,g_1],\alpha_1),c([s_2,g_2],\alpha_2),\ldots \big),
\end{equation}
where $\alpha_t$s and $g_t$s are drawn from the probability distribution $\omega(g_{t+1},\alpha_t \mid g_t,o_t)$. In this setting, Problem~1 can be expressed as
\vspace{0.2cm}
\begin{problem}\textit{
For a given POMDP~$\mathcal{PM}$, a stochastic finite state controller $\mathcal{G}$, a discount factor $\gamma \in (0,1)$, and a total risk  functional $V_\gamma$ as in~\eqref{eq:objriskfsc} with $\{\rho_t\}_{t=1}^\infty$ being coherent risk measures, compute 
$$
(\omega^*,\kappa^*) \in \mathrm{argmin}_{\omega,\kappa} V_\gamma(\mathcal{G},\iota_{\mathrm{init}}).
$$
}
\end{problem}
\vspace{0.2cm}
{The optimal value of Problem~2 provides an upper-bound to that of Problem~1, since a stochastic finite state controller only contains finite memory states and it can be at best as good as the belief-based optimal policy (with infinite memory). The latter claim can also be shown using~\cite[Theorem 1]{hansen1998solving}, which indicates that any improvement in the parameters of a stochastic finite state controller (in the sense of optimizing the value functions) is at most as good as the belief value function.}

For POMDPs controlled by stochastic finite state controllers, the dynamic program is developed in the global state space
$\mathcal{S}\times G$. The value function is defined over this global state space, and policy
iteration techniques must also be carried out in the global state space. For a given stochastic finite state controller, $\mathcal{G}$, and the POMDP $\mathcal{PM}$, the value function $V_{\gamma,\mathcal{M}}([s,g])$ is
the discounted dynamic risk measure under $\mathcal{G}$, and can be computed by solving a set of
 equations:
\begin{multline}\label{eq:valueiterationsfc}
    V_{\gamma,\mathcal{M}}([s,g])\\ = \sum_{\alpha \in Act}   {\sum_{g' \in \mathcal{G}, o \in \mathcal{O}} \omega(g',\alpha \mid g,o) O(o|g')} c([s,g],\alpha) \\+ \gamma \mathrm{R}\Big \{ V_{\gamma,\mathcal{M}}([s',g']),[s,g], T^{\mathcal{M}}  \left([s',g'] \left|
            [s,g] \right. \right) \Big\},
\end{multline}
where $$p(\alpha \mid g) = {\sum_{g' \in \mathcal{G}, o \in \mathcal{O}} \omega(g',\alpha \mid g,o) O(o|g')}. $$ Then, for each $s$, the optimal value function over the induced Markov Chain $\mathcal{M}$ can be computed by taking the minimum of the above equation over all I-states $g$
\begin{equation}\label{eq:dsdsdscmk}
        V^*_{\gamma,\mathcal{M}}(s) := \min_{g \in \mathcal{G}} V_{\gamma,\mathcal{M}}([s,g]).
\end{equation}
 Since $v \mapsto \mathrm{R}(v,\cdot,\cdot)$ is convex (because $\mathrm{R}$ is a coherent risk measure), \eqref{eq:valueiterationsfc}  can be solved by a convex optimization. 

We end this section by demonstrating that the optimal values obtained using the stochastic finite state controllers upper-bound those of the belief-based (infinite-memory) policy. 
\vspace{0.2cm}
\begin{proposition}\textit{
Consider the POMDP $\mathcal{PM}$ and the Markov chain $\mathcal{M}$ induced by the stochastic finite state controller $\mathcal{G}$. Then, for all $s \in \mathcal{S}$, we have $V^*_\gamma(b(s)) \le V^*_\mathcal{\gamma,M}(s)$.}
\end{proposition}
\vspace{0.2cm}
\begin{proof}
The value function of the induced Markov chain $\mathcal{M}$ satisfies~\eqref{eq:valueiterationsfc} for all $[s,g] \in \mathcal{S}\times \mathcal{G}$. For each I-state $g$, the value function in beliefs can be computed as
$$
V_\gamma([b,g]) := \sum_{s \in \mathcal{S}} b(s) V_{\gamma,\mathcal{M}}([s,g]),
$$
and the optimal value function given by
$$
V^*_\gamma(b) = \min_{g \in \mathcal{G}} \sum_{s \in \mathcal{S}} b(s) V_{\gamma,\mathcal{M}}([s,g]).
$$
Applying H\"older inequality to the right-hand side of the above equality, we obtain
\begin{align*}
V_\gamma^*(b) &\le \min_{g \in G} \left(\sup|\sum_{s\in \mathcal{S}} b(s) | \right) |V_{\gamma,\mathcal{M}}([s,g])| \\
&= \min_{g \in \mathcal{G}} V_{\gamma,\mathcal{M}}([s,g]),
\end{align*}
where in the last inequality we used the fact that $\sum_{s}b(s)=1$ since $b \in \Delta(\mathcal{S})$ and the fact that $V_{\gamma,\mathcal{M}}([s,g])$ is non-negative (since $c$ is non-negative). From~\eqref{eq:dsdsdscmk}, we infer $V^*_\gamma \le V^*_{\gamma,\mathcal{M}}$.
\end{proof}

\section{A Bounded Policy Iteration Algorithm for Risk-Averse stochastic finite state controllers} \label{sec:bpi}

So far, we showed that  synthesizing an infinite memory controller for POMDPs with coherent risk objectives is undecidable. On the other hand, a stochastic finite state controller can upper-bound the coherent risk  for a POMDP. In this section, we provide a computational method based on bounded policy iteration to design risk-averse stochastic finite state controllers. Furthermore, we propose techniques for minimizing the upper bound on the total coherent risk by adding I-states to the algorithm in order to escape local minima.


Policy iteration incrementally improves a controller by alternating between two steps: Policy Evaluation and Policy Improvement, until convergence to an optimal policy~\cite{bertsekas76}.  During policy improvement, a dynamic programming update using the so called \emph{dynamic programming backup equation~(DP Backup)}  is used. For a risk-averse POMDP, the DP Backup is given by
\begin{multline*}
V_{\gamma}(b) =  \min_{\alpha \in Act}\Big( c(b,\alpha) +\gamma \mathrm{R}\big \{ V_\gamma(b), b, p(b'\mid b, \alpha) \big\}  \Big),
\end{multline*}
The r.h.s. of the DP Backup  can be applied to any risk value function. The effect is a risk reduction (if possible) at every belief state.  
However, DP Backup is difficult to use directly as it must be computed at each belief state in the belief space, which is uncountably infinite.


In \cite{PoupartB03,hansen08}, a methodology called the Bounded Policy Iteration is proposed for stochastic finite state controllers, which allows stochastic finite state controllers with fewer I-states to have comparable performance in comparison with deterministic finite state controllers, while allowing the stochastic finite state controller to grow in a bounded fashion -- only one (or a few) I-state(s) need to be added at a time to escape a local minima. 

Before  presenting our proposed bounded policy iteration method for risk-averse stochastic finite state controllers, we recall the following important definition.



\vspace{0.2cm}
\begin{defn}[Tangent Belief State]{
A belief state $b$ is called a \emph{tangent belief state}, if $V_\gamma(b)$ touches the DP Backup of
$V_\gamma(b)$ from above. Since $V_\gamma(b)$ must equal $V_{g}^{\beta}$ for some $g$, we also say
that the I-state $g$ is tangent to the backed up value function $V_\gamma$ at $b$.}
\end{defn}
\vspace{0.2cm}
\noindent
Equipped with this definition, the two steps involved in our algorithm is described next.

\subsection{I-States Improvement via  Convex Optimization}
Let $\vec{ V}_{\gamma,\mathcal{M}}(g) \in \mathbb{R}^{|S|}$ denote the vectorized $ V_{\mathcal{M}}([s,g])$ in $s$. We say that an I-state $g$ is  \emph{improved}, if the tunable stochastic finite state controller parameters associated with that I-state can be adjusted so that $\vec{ V}^*_{\gamma,\mathcal{M}}(g)$ decreases. 

As a first step, we point out that the search over $\kappa$ can be dropped. This is simply because the initial I-state is chosen by computing the best valued I-state for the given initial belief, \textit{i.e.}, $\kappa(g_{\mathrm{init}})  =  1$, where $$g_{\mathrm{init}}  =  \underset{g}{\mbox{argmin}}~ \left(\vec{\iota}_{\mathrm{init}}^{\mathcal{M}}\right)^T\vec{V}_{\gamma,\mathcal{M}}(g).$$

After initialization, we pose the improvement as a convex optimization as follows:

\noindent
{\bf I-state Improvement Convex Optimization:}
For the I-state $g$, the following convex optimization is constructed over the variables $\epsilon$, $\omega(g',\alpha|g,o)$, $\forall g',\alpha,o$
\begin{eqnarray}\label{eq:istateimprovmentCOpt}
    &\underset{\epsilon> 0,\omega(g',\alpha|g,o)}{\max} \ \ \ \epsilon& \nonumber \\
    &{\text{subject to}}&  \nonumber \\
    &{\text{Improvement Constraint:}}&  \nonumber\\
    &V_{\gamma,\mathcal{M}}([s,g]) + \epsilon  \le  \text{r.h.s. of \eqref{eq:valueiterationsfc}},~~ \forall s \in \mathcal{S}, & \nonumber \\
&{\text{Probability Constraints:}}&  \nonumber\\
&\underset{(g',\alpha)\in G\times Act}{\sum}\omega(g',\alpha\mid g,o)=1,~~ \forall o \in \mathcal{O},& \nonumber \\
&\omega(g',\alpha \mid g, o)\ge 0,~~\forall g'\in G, \alpha \in Act, o \in \mathcal{O}.&
\end{eqnarray}

The above convex optimization searches for $\omega$ values that improve the I-state value vector $\vec{ V}^*_{\gamma,\mathcal{M}}(g)$ by maximizing the decision variable $\epsilon$. If an improvement is found, \textit{i.e.}, $\epsilon > 0$, the parameters of the I-state are updated by the corresponding minimizing $\omega$. 

Algorithm \ref{algo:policyiteration} outlines the main steps in the bounded policy iteration for risk-averse stochastic finite state controllers. The algorithm has two distinct parts. First, for fixed  parameters of the stochastic finite state controller ($\omega$), policy evaluation is carried out, in which $V_{\gamma,\mathcal{M}}([s,g])$ is computed using the following convex optimization (Steps 2, 10 and 18): For each I-state $g$, we have the following: 
\begin{eqnarray}\label{eq:policyevaluationCP}
    &\underset{\epsilon_1> 0,\epsilon_2> 0,V_{\gamma,\mathcal{M}}}{\min} \ \ \ \epsilon_1-\epsilon_2& \nonumber \\
    &{\text{subject to}}&  \nonumber \\
    &V_{\gamma,\mathcal{M}}([s,g]) - \big(\text{r.h.s. of \eqref{eq:valueiterationsfc}}\big)    \le  \epsilon_1,~~ \forall s \in \mathcal{S}, & \nonumber \\
        &V_{\gamma,\mathcal{M}}([s,g]) - \big(\text{r.h.s. of \eqref{eq:valueiterationsfc}}\big)    \ge  \epsilon_2,~~ \forall s \in \mathcal{S}.
\end{eqnarray}
In fact, the above optimization solves~\eqref{eq:valueiterationsfc} for  $V_{\gamma,\mathcal{M}}$.
 Second, after evaluating the current coherent risk function, an improvement is carried out either by changing the parameters of existing I-states, or if no new parameters can improve any I-state, then a fixed number of I-states are added to escape the local minima (Steps 14-17). This is described in Section~\ref{sec:escape}.

\begin{algorithm}
\caption{Bounded Policy Iteration For Synthesizing Risk-Averse Stochastic Finite State Controllers}
\label{algo:policyiteration}
\begin{algorithmic}[1]
\REQUIRE (a) An initial feasible stochastic finite state controller, $\mathcal{G}$. (b) Maximum size of stochastic finite state controller $N_{max}$. (c) $N_{new} \le N_{max}$ number of I-states
\STATE $improved \leftarrow True$
\STATE Compute the value vectors, $\vec{V}_{\gamma,\mathcal{M}}$ based on the convex optimization~\eqref{eq:policyevaluationCP}.
\WHILE {$|G| \le N_{max}$ \AND $improved = True$}
\STATE $improved \leftarrow False$
\FORALL {I-states $g \in G$}
\STATE  Set up the I-State Improvement Convex Optimization~\eqref{eq:istateimprovmentCOpt}.
\IF {I-State Improvement Convex Optimization results in optimal $\epsilon > 0$}
\STATE Replace the parameters for I-state $g$
\STATE $improved \leftarrow True$
\STATE Compute the value vectors, $\vec{V}_{\gamma,\mathcal{M}}$ based on the convex optimization~\eqref{eq:policyevaluationCP}.
\ENDIF
\ENDFOR
\IF {$improved = False$ \AND $|G| < N_{max}$}
\STATE $n_{added} \leftarrow 0$
\STATE $N'_{new} \leftarrow \min(N_{new},N_{max}-|G|)$
\STATE {Try to add $N'_{new}$ I-state(s) to $\mathcal{G}$ via Algorithm~2 in Section \ref{sec:escape}.}
\STATE $n_{added} \leftarrow $ actual number of I-states added in previous step.
\IF {$n_{added} > 0$}
\STATE $improved \leftarrow True$
\STATE Compute the value vectors, $\vec{V}_{\gamma,\mathcal{M}}$ based on the convex optimization~\eqref{eq:policyevaluationCP}.
\ENDIF
\ENDIF
\ENDWHILE
\ENSURE $\mathcal{G}$
\end{algorithmic}
\end{algorithm}

\subsection{Escaping Local Minima by Adding I-States}
\label{sec:escape}
At some point of running the algorithm,  no I-state may be improved with further iterations, \textit{i.e.}, $\forall g\in G$, the corresponding convex optimization~\eqref{eq:istateimprovmentCOpt} yields an optimal value of $\epsilon = 0$. Then, policy iteration has reached a local minimum if and only if $\vec{ V}_{\gamma,\mathcal{M}}(g)$ is tangent to the backed up value function for all $g\in G$~\cite{PoupartB03}. The dual variables corresponding to the Improvement Constraints in  \eqref{eq:istateimprovmentCOpt} provide those belief states that are tangent to the  risk function. The process for adding I-states  involves forwarding the tangent beliefs one step and then checking if the value of those forwarded beliefs can be improved. The procedure for adding I-states is provided in Algorithm \ref{algo:addnode}.

\begin{algorithm}
\caption{{Adding I-states to Escape Local Minima}}
\label{algo:addnode}
\begin{algorithmic}[1]
\REQUIRE (a) Set $B$ of tangent beliefs for each I-state. (b) A function $node:B\to G$ identifying the I-state which yields each tangent belief. (c) $N_{new}$ the maximum number of I-states to add. (d) $\vec{V}_{\gamma,\mathcal{M}}(g)$ the computed risk value functions at each node $g \in G$.
\STATE $N_{added} \leftarrow 0$.
\REPEAT {}
  \STATE Pick $b\in B$, $B \leftarrow B\backslash\{b\}, g \leftarrow node(b)$.
   \STATE $Fwd = \emptyset$
  \FORALL {$(\alpha,o)\in (Act\times \mathcal{O})$}
    \IF { $Pr(o|b) = \sum_{s\in\mathcal{S}}b(s)O(o|s) > 0$ }
      \STATE  Look ahead one step to compute forwarded beliefs 
$
          b_{o,\alpha}(s') = \sum_{s}T(s'|s,\alpha)\frac{O(o|s)b(s)}{\sum_{o'\in\mathcal{O}}O(o'|s)b(s)}.
$
      \STATE $Fwd \leftarrow Fwd \cup \{b_{o,\alpha}\}$.
    \ENDIF
  \ENDFOR
  \FORALL {$b\in Fwd$}
    \STATE {Apply a dynamic programming backup step
\begin{multline*}
V_{\gamma}^{BU}(b) =  \min_{\alpha \in Act}\Big( c(b,\alpha) \\+\gamma \mathrm{R}\big \{ V_\gamma(b), b, p(b'\mid b, \alpha) \big\}  \Big),
\end{multline*}
    where $V_\gamma(b(s))=\min_{g\in G} b_{o,\alpha}(s)V_{\gamma,\mathcal{M}}([s, g])$ and $b_{o,\alpha}$ is computed for each product state $s'\in \mathcal{S}$ as follows
$
b_{o,\alpha}(s') = \sum_{s}T(s'|s,\alpha)\frac{O(o|s)b(s)}{\sum_{o'\in\mathcal{O}}O(o'|s)b(s)}.
$}
    \STATE Note the minimizing action $\alpha^*$ and I-state $g^*$.
  \ENDFOR
  \IF{$V^{BU}_{\gamma}(b) < V_{\gamma}(b)$ for $b \in Fwd$}
      \STATE Add new deterministic I-state $g_{new}$ such that $\omega(g_{new}|g^*,\alpha^*,o) = 1$, $\forall o\in\mathcal{O}$.
      \STATE $N_{added} \leftarrow N_{added} + 1$.
  \ENDIF
  \IF  {$N_{added} \ge N_{new}$}
    \RETURN
  \ENDIF
\UNTIL $B = \emptyset$.
\end{algorithmic}
\end{algorithm}

Algorithm \ref{algo:addnode} can be understood as follows. Assume that
a tangent belief $b$ exists for some I-state $g$. Instead of directly improving the value of the
tangent belief, the algorithm tries to improve the value of forwarded
beliefs reachable in one step from the tangent beliefs. First, the forwarded beliefs are computed (Step 4-8). Then, the corresponding  risk value functions are applied to a DP Backup~(Steps 9-11). If some action $\alpha^*$
and successor I-state $g^*$ can in fact reduce the risk value (Step 12), then a new
I-state is added which deterministically leads to this action and
successor I-state~(Steps 13-14). Note that at the end of the algorithm,
the newly added I-states, $g_{new}$ have no incoming
edges, \textit{i.e.}, no pre-existing I-states transition to
$g_{new}$. However, when the other I-states are improved in subsequent
policy improvement steps, they generate transitions to any $g_{new}$
added. This new I-state then improves the value of the original tangent belief.

\section{Numerical Example}\label{sec:example}

An  agent (e.g. a robot) has to autonomously navigate a two dimensional terrain map (e.g. Mars surface) represented by a $10\times 10$ grid world ($100$ states) with $15$ obstacles of different shapes. At each time step the agent can move to any of its eight neighboring states (diagonal moves are allowed). Due to sensing and control noise, however, with probability $\delta$ a move to a random neighboring state occurs. The stage-wise cost of
each move until reaching the destination is $1$, to account for fuel usage. In between the
starting point and the destination, there are a number of obstacles that the agent should
avoid. Hitting an obstacle incurs the cost of $ 10$ leading to termination, while the goal grid region has reward $80$. The discount factor is $\gamma=0.95$. After a move is chosen, the observation of  the agent is assumed to be binary, \textit{i.e.}, either an obstacle is detected in the next cell that the robot is moving to or not. Similar to~\cite{chow2015risk}, in our simulations, we included an
obstacle and target position perturbation in a random direction to one of the neighboring grid
cells with probability $0.2$ to represent uncertainty in the terrain map~(recall the uncertainty in Mars terrain maps as shown in Figure~1).
 
The objective is to compute a safe (\textit{i.e.}, obstacle-free) path that is fuel efficient. To this end, we consider CVaR as the coherent risk measure. CVaR is given by 
\begin{equation}
    \rho_t(c_{t+1}) = \inf_{z \in \mathbb{R}} \left\{ z + \frac{1}{\alpha} \mathbb{E}\left[  (c_{t+1}-z)_{+} \mid \mathcal{F}_t    \right]                        \right\},
\end{equation}
where $(\cdot)_{+}=\max\{\cdot, 0\}$ and the infimum should be understood point-wise. In general, the confidence level $\alpha$ may be $\mathcal{F}_t$-measurable function with values in the interval $(0,1)$. Here, we assume  $\alpha \in (0,1)$. A value of $\alpha \simeq 1$ corresponds to a risk-neutral policy; whereas, a value of $\alpha \simeq 0$ is rather a risk-averse policy. For CVaR risk measure, \eqref{eq:valueiterationsfc} can be computed as
\begin{align*}
\hspace{-.5cm}    V_{\gamma,\mathcal{M}}([s,g]) =& \sum_{\alpha , g' , o }   \omega(g',\alpha \mid g,o) O(o|g') c([s,g],\alpha) \\&+ \gamma\inf_{z \in \mathbb{R}} \bigg\{ z + \frac{1}{\alpha} \sum_{g',s', o,\alpha} \left( V\left([s',g']\right)-z\right)_{+} \\
    &~~\times O(o\mid s)\omega(g',\alpha \mid g, o) T(s'\mid s,\alpha)\bigg\},
\end{align*}
where the infimum on the right hand side of the above equation can either be solved by line search techniques or
by representation in terms of an elementary linear programming problem since it is convex in $z$~\cite[Theorem 1]{rockafellar2000optimization} (the function $(\cdot)_+$ is increasing and convex~\cite[Lemma A.1., p. 117]{ott2010markov}).

 \begin{figure}[t] \centering{
 \hspace*{-.8cm}\includegraphics[scale=.31]{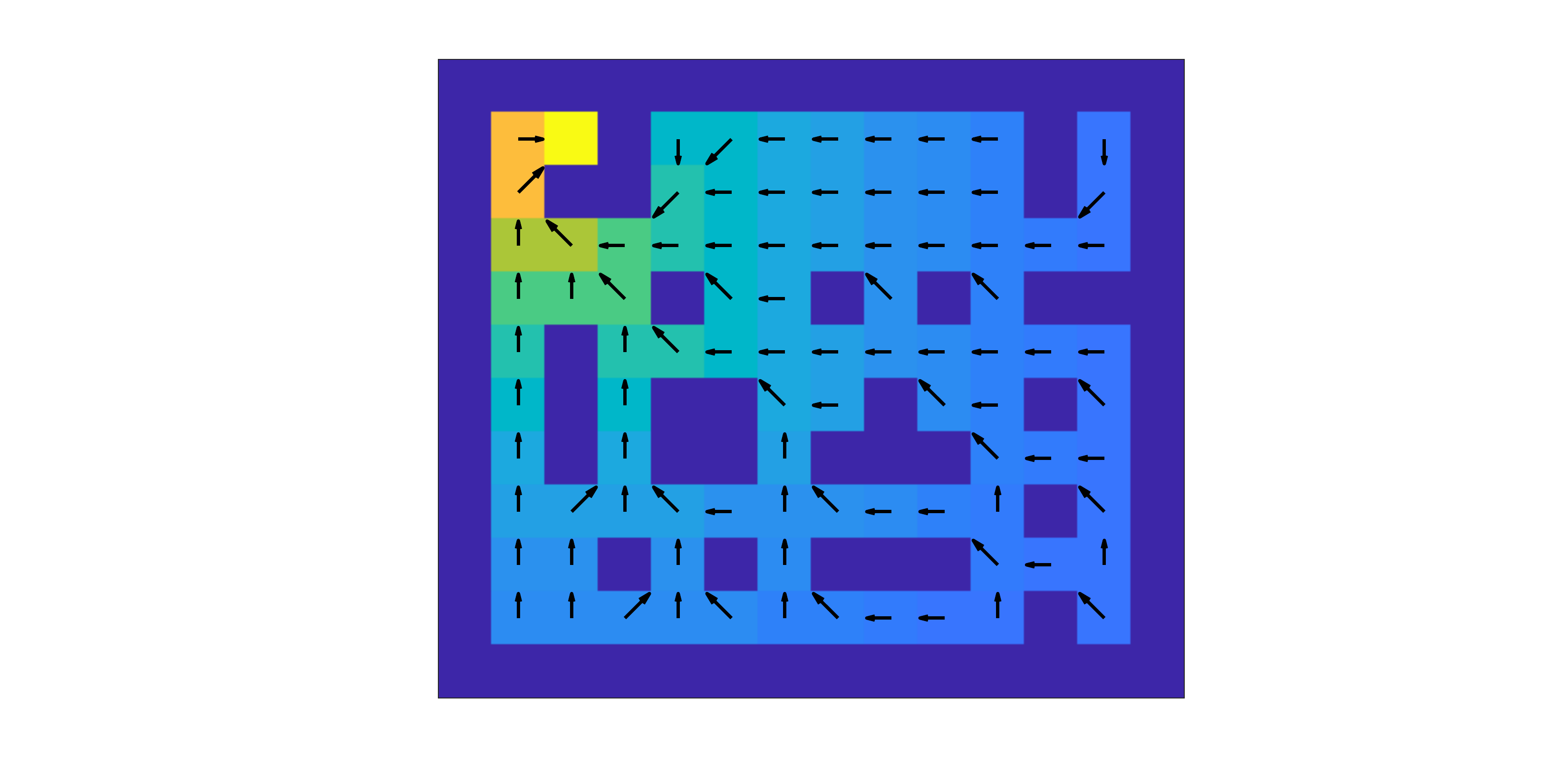}~~~~~~~~\\
\vspace{-.6cm}
 \hspace*{-.8cm}\includegraphics[scale=.31]{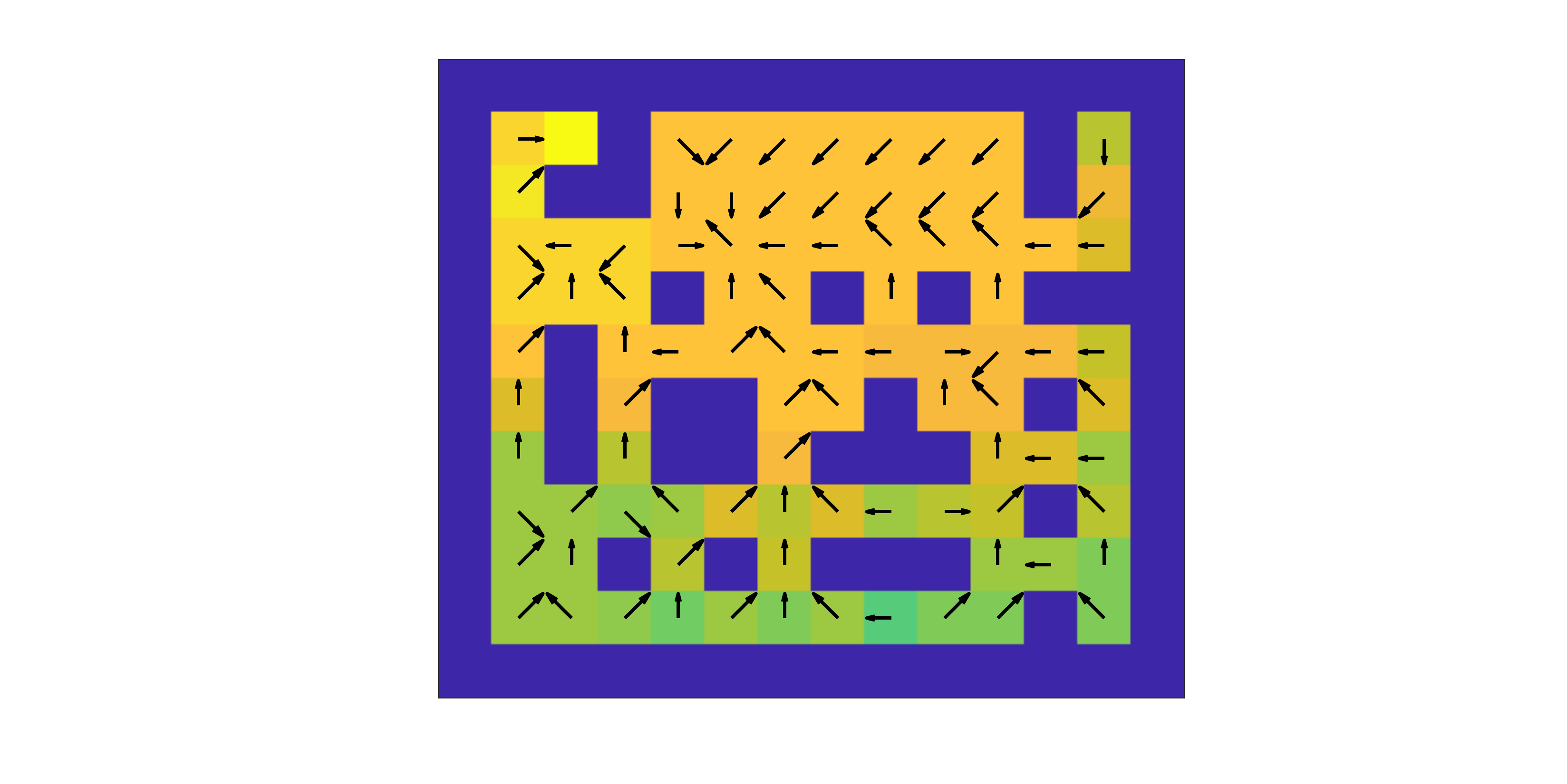}
\vspace{-1cm}
\caption{Numerical results obtained based on the proposed risk-averse control method for two different confidence levels of (top) $\alpha=0.9$ and (bottom) $\alpha=0.1$. The yellow square at $(1,2)$ denotes the goal region. The arrows represent the actions (or the moves) with the highest probability.
}} \label{fig:result}
 \end{figure}

Figure~2 depicts the policies and the value functions computed for the grid world based on the bounded policy iteration technique in Section~\ref{sec:bpi}. For these experiments, we used $2$ internal states for the stochastic finite state controller and the corresponding convex optimizations were solved using CVX toolbox~\cite{cvx} in MATLAB.

 As it can be observed from Figure~2,    the risk-neutral policy
leads to shorter paths from different cells to the target. However,  on $100$ perturbed scenarios, it performed poorly with $43$ failures. On the other hand,
the risk-averse policy leads to longer routes from cells to the target chooses, but it resulted only in $3$ failed scenarios. These results parallel those obtained in~\cite{chow2015risk}, wherein  risk-averse policies in terms of CVaR for MDPs were studied.

\section{Conclusions} \label{sec:conclusions}

We proposed a method based on bounded policy iteration and convex optimization to design risk-averse stochastic finite state controllers for POMDPs. Future research will explore risk-averse polices for POMDPs that maximize the  satisfaction probability of a set of high-level mission specifications in terms of temporal logic formulae~\cite{sharan14,MSB19}. Furthermore, the risk-averse policy synthesis technique will be applied for designing risk-averse planning policies for traversing on uncertain Mars surface (as depicted in Figure~1).

\footnotesize{
\bibliography{main}
}
\bibliographystyle{plain}

\end{document}